\documentclass[sigconf]{acmart}
\AtBeginDocument{%
  }

\setcopyright{acmlicensed}
\copyrightyear{2026}
\acmYear{2026}

\setcopyright{cc}
\setcctype{by}
\acmConference[SIGIR '26] {Proceedings of the 49th International ACM SIGIR Conference on Research and Development in Information Retrieval}{July 20--24, 2026}{Melbourne, VIC, Australia.}
\acmBooktitle{Proceedings of the 49th International ACM SIGIR Conference on Research and Development in Information Retrieval (SIGIR '26), July 20--24, 2026, Melbourne, VIC, Australia}
\acmDOI{3805712.3809661}
\acmISBN{979-8-4007-2599-9/2026/07}

\usepackage{enumitem}
\usepackage{multirow}
\usepackage{array} 
\usepackage{boldline} 
\usepackage{booktabs}
\usepackage{xspace}
\usepackage{array}
\usepackage{graphicx}
\usepackage{subcaption}
\usepackage{graphicx}

\newcommand{\ProjectName}[1]{{\small\textsc{QGA}}} 

\newcommand{\eat}[1]{}
\newcommand{\paratitle}[1]{\vspace{1.5ex}\noindent \textbf{#1}}
\newcommand{\ie}{\emph{i.e.,}\xspace}

\newcommand{\baby}{\textsc{Guide}\xspace}

\begin{document}

\title{Generative Auto-Bidding with Unified Modeling and Exploration}

\author{Mingming Zhang}
\authornote{Work done when Mingming Zhang and Na Li were interns at Taobao \& Tmall Group of Alibaba.}
\affiliation{%
  \institution{Key Laboratory of Aerospace Information Security and Trusted Computing, Ministry of Education, School of Cyber Science and Engineering, Wuhan University\\Taobao \& Tmall Group of Alibaba}
  \city{Wuhan}
  \country{China}
}
\email{mingmingzhang@whu.edu.cn}

\author{Feiqing Zhuang}
\authornotemark[2]
\affiliation{%
  \institution{Taobao \& Tmall Group of Alibaba}
  \city{Hangzhou}
  \country{China}}
\email{zhuangfeiqing.zfq@alibaba-inc.com}

\author{Na Li}
\authornotemark[1]
\affiliation{%
  \institution{Key Laboratory of Aerospace Information Security and Trusted Computing, Ministry of Education, School of Cyber Science and Engineering, Wuhan University\\Taobao \& Tmall Group of Alibaba}
  \city{Wuhan}
  \country{China}
}
\email{wannal@whu.edu.cn}

\author{Shengjie Sun}
\affiliation{%
  \institution{Taobao \& Tmall Group of Alibaba}
  \city{Hangzhou}
  \country{China}}
\email{shengjie.ssj@alibaba-inc.com  }

\author{Xiaowei Chen}
\affiliation{%
  \institution{Taobao \& Tmall Group of Alibaba}
  \city{Hangzhou}
  \country{China}}
\email{qisheng.cxw@alibaba-inc.com}

\author{Junxiong Zhu}
\affiliation{%
  \institution{Taobao \& Tmall Group of Alibaba}
  \city{Hangzhou}
  \country{China}}
\email{xike.zjx@alibaba-inc.com}

\author{Fei Xiao}
\affiliation{%
  \institution{Taobao \& Tmall Group of Alibaba}
  \city{Hangzhou}
  \country{China}}
\email{guren.xf@alibaba-inc.com}

\author{Keping Yang}
\affiliation{%
  \institution{Taobao \& Tmall Group of Alibaba}
  \city{Hangzhou}
  \country{China}}
\email{shaoyao@alibaba-inc.com}

\author{Lixin Zou}
\affiliation{%
  \institution{Key Laboratory of Aerospace Information Security and Trusted Computing, Ministry of Education, School of Cyber Science and Engineering, Wuhan University}
  \city{Wuhan}
  \country{China}
}
\email{zoulixin@whu.edu.cn}

\author{Chenliang Li}
\authornote{Corresponding authors.}
\affiliation{%
  \institution{Key Laboratory of Aerospace Information Security and Trusted Computing, Ministry of Education, School of Cyber Science and Engineering, Wuhan University}
  \city{Wuhan}
  \country{China}
}
\email{cllee@whu.edu.cn}

\renewcommand{\shortauthors}{Mingming Zhang et al.}

\begin{abstract}
Automated bidding has become a core component of modern digital advertising. Early methods were primarily rule-based, while easy to implement, they struggled to adapt to rapidly changing environments. Subsequent Reinforcement Learning methods modeled bidding as a Markov Decision Process but were limited in their ability to capture long-term dependencies. While recent generative models have shown encouraging progress, they generally lack explicit mechanisms to balance exploration and safety. They often rely solely on simple action perturbations or trajectory guidance to foster bidding exploration, and critically, they lack a safety fallback mechanism. This limitation leads to inefficient exploration and significantly increases the financial risk for advertising platforms.

To bridge this gap, we propose a new framework named \textbf{G}enerative Auto-Bidding with \textbf{U}n\textbf{i}fied Mo\textbf{d}eling and \textbf{E}xploration (\baby), which synergistically integrates directed exploration with a safe fallback mechanism. \baby utilizes a Decision Transformer (DT) to jointly model historical bidding actions and environmental state transitions. A Q-value module guides the DT's exploration through regularization constraints. Concurrently, an Inverse Dynamics Module (IDM) leverages the future states predicted by the DT to infer robust and behaviorally consistent actions, thereby providing a safe policy fallback. The Q-value module then adaptively selects the final action from these two options, balancing exploration and safety. Together, these three components form an integrated "explore--safeguard--select" pipeline, unifying efficiency and safety.

We conduct comprehensive experiments on public datasets, in simulated auction environments, and through a large-scale online deployment on Taobao, a leading advertising platform in China. The results demonstrate that \baby consistently outperforms state-of-the-art (SOTA) baseline methods across all scenarios. In real-world online deployment, \baby achieves remarkable improvements: +4.10\% in ad GMV, +1.40\% in ad clicks, +1.66\% in ad cost, and +3.52\% in ad ROI, demonstrating its effectiveness and strong industrial applicability.
\end{abstract}

\ccsdesc[500]{Applied computing~Online auctions}
\ccsdesc[500]{Information systems~Computational advertising}

\keywords{Auto Bidding, Generative Decision Model, Decision Transformer}

\maketitle

\section{Introduction}
With the rapid evolution of the digital advertising ecosystem, the global online advertising market has already reached the hundred-billion-dollar scale in 2025, traditional manual Ad bidding methods struggle to meet the demands for real-time response and large-scale optimization \cite{borissov2010automated, wen2022cooperative}. Automated bidding technology not only enhances the efficiency of ad delivery but also allows for more precise budget allocation and resource management according to different marketing objectives, such as clicks, conversions, or return on investment \cite{zhang2014optimal,li2024trajectory,liu2020dynamic,yuan2022actor,zhang2023personalized,yuan2013real, li2022auto}. Its growing importance in improving advertising effectiveness and reducing operational costs has made it one of the core tools in contemporary advertising strategies.
\begin{figure}
    \centering
    \includegraphics[width=0.9\linewidth]{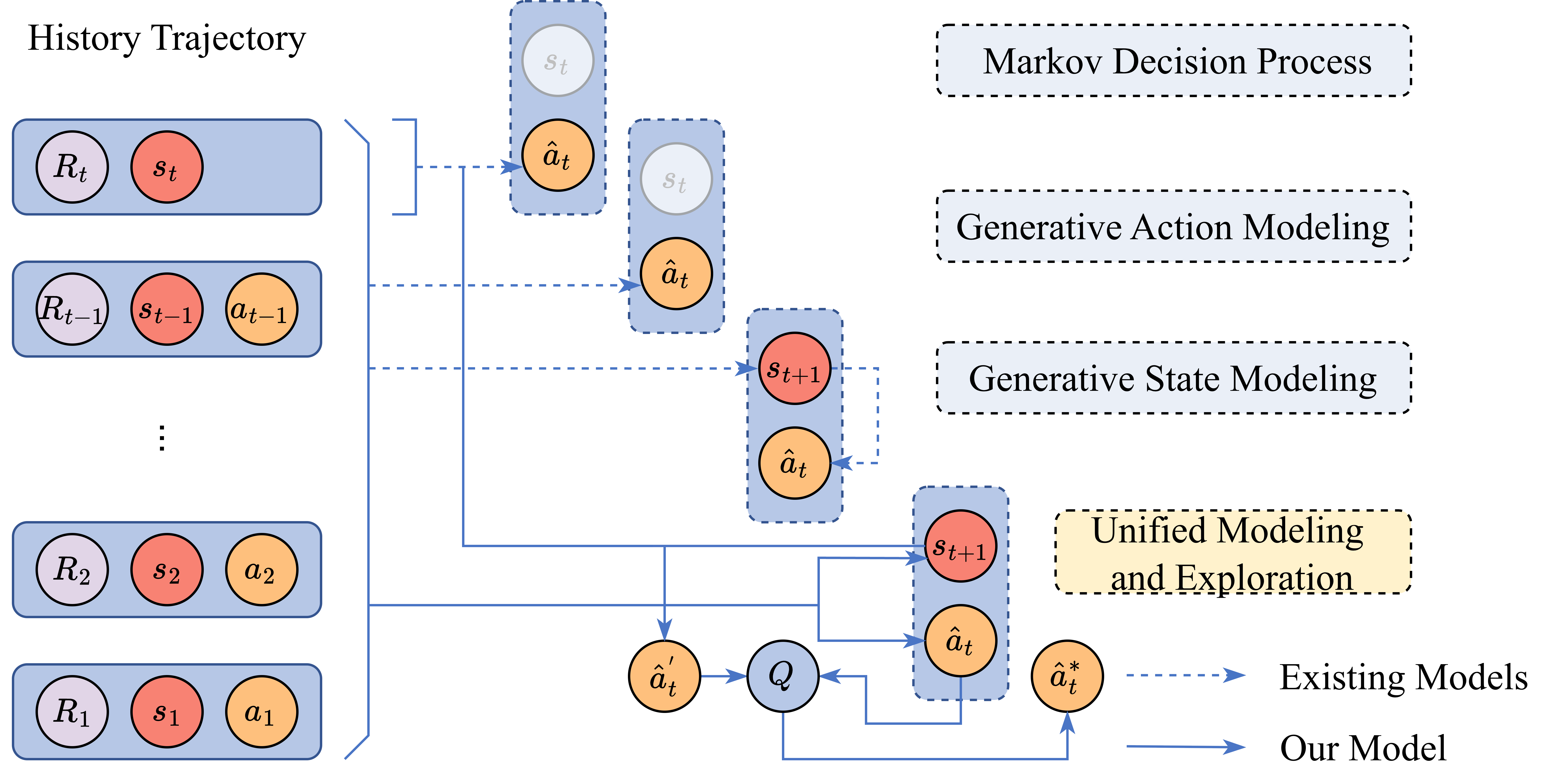}
    \caption{Different Modeling Approaches in Ad Bidding. $a_t$ and $s_t$ denotes the actions and states respectively, and $\hat a_t^*$ denotes the better action. Q donates Q-value Module.}
    \label{fig:overalCmp}
    \vspace{-0.25cm}
\end{figure}

Early automated bidding approaches often relied on rule-based strategies, such as PID control. Although these methods are easy to implement, they lack the ability to adapt to dynamic advertising environments. To address these limitations, reinforcement learning has been widely applied to automated bidding tasks, modeling them as a Markov Decision Process (MDP) \cite{puterman2014markov, boutilier1999decision}, where the advertiser’s bidding behavior in each auction is treated as a decision action, and optimal actions are chosen based on the environmental state, such as user characteristics, accumulated rewards, and market dynamics~\cite{fujimoto2019benchmarking, kostrikov2021offline}. However, since MDPs consider only the current state and action, they often fail to adequately capture complex temporal dependencies and dynamics of the advertising environment, making it difficult to make accurate decisions when faced with long-term dependencies and complex user behavior patterns~\cite{cai2017real}.

Recently, generative models have demonstrated the ability to effectively model complex historical dependencies and are able to discover improved bidding strategies, making them a focal point of current research~\cite{jiang2025optimal, gao2025generative, li2025gas, guo2024generative}. Models based on Decision Transformer (DT) \cite{chen2021decision}, such as GAS \cite{li2025gas} and GAVE \cite{gao2025generative}, model the sequence of bidding actions, while those based on Decision Diffusion (DD) \cite{lu2025makes} , such as AIGB \cite{guo2024generative} and EGDB \cite{peng2025expert}, model the sequence of advertising environment states. These models have achieved impressive results by integrating thoughtfully designed exploration strategies.

However, underlying these advancements lies a fundamental challenge: how to encourage exploration while ensuring the financial safety of the platform. It is well known that there is an inherent tension between exploration and reliability. In the high-stakes environment of ad auctions, where every cent is critical, unconstrained exploration is tantamount to gambling. While existing approaches promote exploration through techniques such as action perturbation and value guidance, they generally lack an explicit safety fallback mechanism. When the model explores into unknown or perilous policy spaces, the system is unable to revert to a known, robust baseline policy, rendering the exploration process not only inefficient but also exceptionally risky. This leads to a critical open question: \textit{how can we design a unified framework that synergistically integrates directed, efficient exploration with a robust safety fallback mechanism, thereby achieving both high performance and operational reliability?}

To address this challenge, we propose \textbf{G}enerative Auto-Bidding with \textbf{U}n\textbf{i}fied Mo\textbf{d}eling and \textbf{E}xploration (denoted as \baby), a unified framework that integrates exploration effectiveness and safety. As illustrated in Figure~\ref{fig:overalCmp}, \baby jointly models environmental dynamics and historical bidding action sequences, complemented by a Q-value-based action optimization and selection module to balance exploration and safety. Specifically, we employ the DT as the backbone network to simultaneously generate future state trajectories and candidate bidding action sequences, thereby gaining a deeper understanding of the current bidding environment. To enable directed exploration, we integrate a Q-value module that guides the exploration direction of the DT through regularization constraints. Meanwhile, we introduce an Inverse Dynamics Module (IDM), which leverages the DT-predicted future states to infer plausible bidding actions from the transition between the current and predicted states. By design, the DT boldly explores potentially high-reward strategies, while the IDM effectively imitates the behavioral policy embedded in the training data, producing safer and more stable actions that serve as a reliable fallback during high-risk exploration. The Q-value module further adaptively selects actions between those proposed by the DT and the IDM, ensuring a balanced trade-off between exploration and safety. These three components work in concert to enable efficient bidding exploration with safety guarantees, leading to smarter and more robust automated bidding. To enable effective model optimization, we also utilize a two-stage training procedure for efficient model learning.

We conduct comprehensive evaluations of \baby on public offline datasets and simulated advertising auction environments, and the results show that \baby significantly outperforms existing state-of-the-art baseline methods in all settings. Furthermore, we deployed \baby on Taobao, one of the largest e-commerce platform in China, and achieve improvement in Ad GMV by
4.10\%, Ad clicks by 1.40\%, Ad cost by 1.66\%, and Ad ROI by 3.52\%, demonstrating the effectiveness and leading performance of \baby in the field of automated bidding. In summary, our contributions are as follows:  
\begin{itemize}[left=0pt]
    \item We propose the first unified modeling paradigm that jointly captures environmental dynamics and bidding actions within a single generative framework, simultaneously modeling the evolution of the advertising environment and the sequence of historical bids. This design significantly enhances understanding of complex, dynamic ad ecosystems for policy optimization.
    \item We propose a novel bidding mechanism that integrates "Exploration–Guarantee–Selection", which effectively resolves the fundamental tension between exploration and safety in high-risk advertising scenarios by organically combining three core components: an active exploration module based on Decision Transformer, a safety fallback module grounded in IDM, and a Q-value-based action selector.
    \item We conduct comprehensive experiments on public offline datasets, simulated scenarios, and real-world business environments. The results show that our proposed \baby significantly outperforms existing state-of-the-art auto-bidding baselines across all metrics and settings.

\end{itemize}

\section{Related Works}
In the domain of online advertising, automated bidding methods have evolved into four main categories: PID control, reinforcement learning, generative models, and LLM-based agents. Early bidding methods, which were based on PID control theory \cite{chen2011real,yang2019bid,zhang2016feedback,knospe2006pid,borase2021review}, suffered from several practical issues, most notably a heavy reliance on meticulous parameter tuning and a limited capacity for adapting to dynamic market environments. To address these inherent limitations, the research community turned its attention to reinforcement learning, giving rise to more advanced bidding algorithms such as USCB \cite{he2021unified} and SORL \cite{mou2022sustainable}. These algorithms leverage fundamental reinforcement learning techniques like IQL \cite{kostrikov2021offline} and CQL \cite{kumar2020conservative} to learn behavioral policies from ad log datasets, enabling fully automated bidding. However, they remain relatively inefficient at fully utilizing the rich historical information in the logs.

Subsequently, generative approaches were introduced, which innovatively reframe the ad bidding task as a sequence generation problem.  Generative efforts can be further divided into two categories: those based on Decision Diffusion \cite{ajay2022conditional, zhu2024madiff} and those based on Decision Transformer \cite{chen2021decision}. AIGB \cite{guo2024generative} pioneered a new generative bidding paradigm, using Decision Diffusion to model ad status sequences and an inverse dynamics model to generate actions. Then EGDB \cite{peng2025expert} introduced expert information to optimize the generated trajectories. GAS \cite{li2025gas} and GAVE \cite{gao2025generative}, on the other hand, are based on Decision Transformer \cite{chen2021decision} networks to generate bidding actions. They designed Monte Carlo post-training search strategy and value-guided action generation optimization to enhance exploration respectively. However, none of the aforementioned methods explicitly consider the trade-off between exploration and safety. With the advancement of LLMs, recent research such as RTBAgent \cite{cai2025rtbagent} has leveraged the planning capabilities of LLM agents to fine-tune the actions of basic bidding models. However, it still faces challenges such as excessive latency and low reproducibility.

Meanwhile, developing high-fidelity advertising simulation environments is widely recognized as crucial for improving automated bidding strategies.
This importance stems from their unique capability to bridge the often-substantial gap between analyses conducted on static offline data and the dynamic, unpredictable nature of real-world auction scenarios \cite{yuan2013real, cai2017real, jin2018real}. AuctionNet \cite{su2024auctionnet} introduced the first large-scale advertising simulation framework with a traffic generator that replicates real-world distributions. More recently, the Bid2X \cite{ji2025bid2x} project advanced this line of research by training the first large-scale, Transformer-based environment model. By leveraging a wide array of diverse advertising datasets, this work not only created a powerful simulator but also made the significant demonstration that its performance scales in accordance with the well-established scaling laws \cite{kaplan2020scaling}, a principle newly verified within the advertising domain.

\section{Preliminary}
\subsection{Definition of Auto-Bidding Problem}

\subsubsection{Problem Setting}

The goal of auto-bidding is to determine a bidding sequence that maximizes the total value of acquired traffic $v_i$, subject to budget and CPA constraints. Let $x_i \in \{0,1\}$ indicate if impression $i$ is won. The optimization problem is:
\begin{equation}
    \max_{\{b_i\}_{i=1}^{I}} \sum_{i=1}^I x_i v_i
\end{equation}
subject to:

\begin{itemize}
    \item \textbf{Budget Constraint:} Total spend should not exceed the budget $B$:
    \begin{equation}
        \sum_{i=1}^I x_i c_i \leq B
    \end{equation}

    \item \textbf{CPA Constraint:} The average cost per acquisition (CPA) should not exceed threshold $C$:
    \begin{equation}
        CPA=\frac{\sum_{i=1}^I x_i c_i}{\sum_{i=1}^I x_i v_i} \leq C
    \end{equation}
\end{itemize}
where $c_i$ is the actual cost incurred for winning the $i$-th impression, and $v_i$ is the value generated by that impression. The budget constraint is strictly enforced, while CPA constraints are typically soft, as they are evaluable only after the campaign ends.

\subsubsection{Optimal Bidding Policy} The optimal bid can then be expressed, via the complementary slackness theorem~\cite{dantzig2016linear}, as a function of its value and the CPA threshold~\cite{yu2017online}:
\begin{equation}
    b_i^* = (\lambda_0^* + \lambda_1^* C)v_i = \lambda^*v_i
\end{equation}
where $\lambda_0^*$, $\lambda_1^*$ are coefficients determined by the campaign's budget and CPA requirements. This formulation allows the bidding strategy to effectively balance value maximization and constraint satisfaction in dynamic auction environments.

\subsection{Sequence Modeling for Auto-bidding Problem}
While the bidding policy $b_i = \lambda^* v_i$ provides a theoretically sound structure, the assumption of a single, static multiplier $\lambda^*$ is insufficient for real-world dynamic auction environments. Market conditions like competitors' bids and impression availability keep changing, so a fixed $\lambda^*$ is no longer optimal over time. A more powerful strategy is to adapt the multiplier at each time step, $\lambda_t$, based on the evolving campaign status and market feedback.

This need for dynamic adjustment naturally casts the auto-bidding problem as a \textbf{sequential decision process}, enabling the application of modern sequence modeling approaches such as the Decision Transformer. This paradigm reformulates the problem as conditional sequence modeling. The goal is to learn a model capable of generating high-return trajectories. Specifically, we aim to model the conditional probability of an action $a_t$, conditioned on the past history of states, actions, and rewards, as well as a desired future performance target.

The main components of this sequential formulation are:

\begin{itemize}
    \item \textbf{State} $s_t$: feature vector summarizing bidding environment at time $t$ (e.g., remaining budget, time, and previous results).
    \item \textbf{Action} $a_t$: adjustable bidding parameter for time $t$ (e.g., $a_t = \lambda_t$).
    \item \textbf{Reward} $r_t$: total value from impressions won in $t$ for $N_t$ candidate impressions, $r_t = \sum_{n=1}^{N_t} x_{n} v_{n}$.
    \item \textbf{Return-to-go} $R_t$: cumulative rewards from $t$ to $T$, $R_t = \sum_{t' = t}^T r_{t'}$.
\end{itemize}

The bidding process forms a trajectory $\tau = (s_1, a_1, r_1, \ldots, s_T, a_T, r_T)$, suitable for sequence modeling approaches such as transformer or diffusion architectures. This enables flexible and adaptive policy optimization in dynamic ad environments.

\section{Method}
\begin{figure*}[htbp]
    \includegraphics[width=0.9\linewidth]{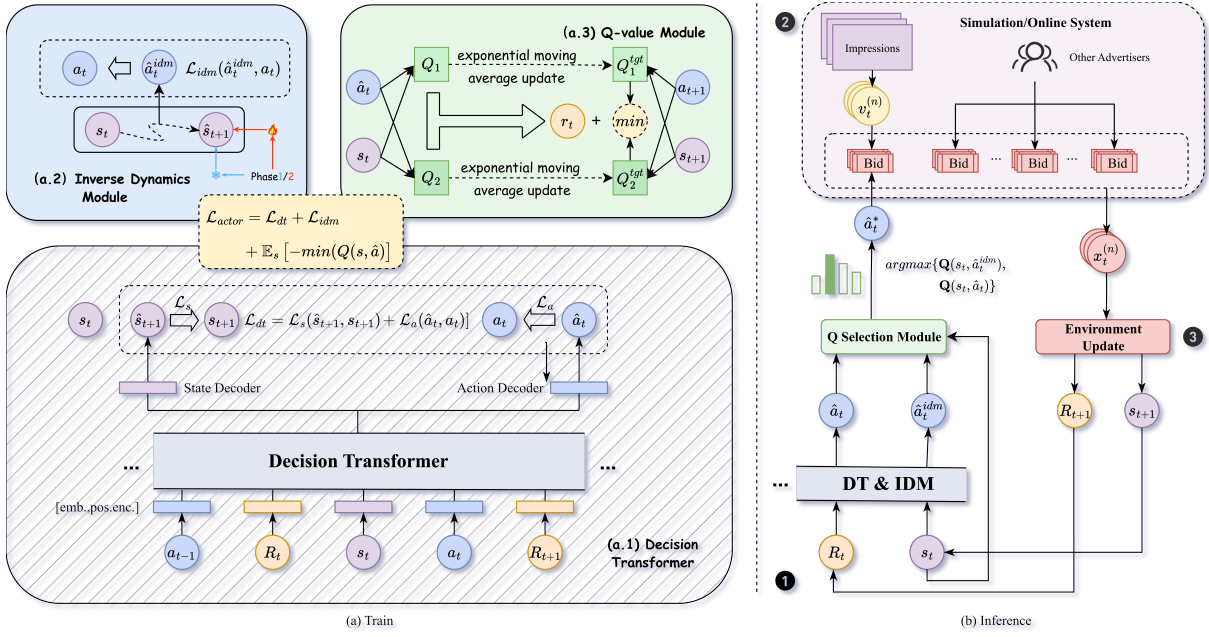}
    \caption{Overview architecture. a) Training of the unified modeling framework. b) Inference with bid selection}
    \label{fig:overview}
\end{figure*}

Here, we first introduce the design details for the two basic modules of our proposed \baby: DT and IDM as well as the two-stage training procedure. Then, we present the design of the Q-value module. Next, we describe the mechanism for action selection using the Q-value module. Finally, we make a summary towards the role of DT and IDM respectively. The overall architecture is shown in Figure \ref{fig:overview}. 

\subsection{Unified Modeling of Bid Trajectories}

\subsubsection{Trajectory Construction and Modeling}
In the auto-bidding task, each round of bidding can be represented as a temporal trajectory that sequentially records the advertising environment states, bidding actions, and the resulting rewards. Formally, a trajectory can be represented as follows:
\begin{equation}
    \tau = (s_1, a_1, r_1, s_2, a_2, r_2, ..., s_T, a_T, r_T)
\end{equation}
where $s_t$ denotes the environment state at time step $t$, $a_t$ and $r_t$ are the bidding action taken and  the immediate reward received respectively at the same step.

To effectively capture historic information and long-term dependencies, we adopt the Decision Transformer for historical sequence modeling. At each time step, the DT module takes in the historical states, actions, and return-to-go as input features, and predicts the next action and next state in one go. Specifically, at time $t$, the model makes the prediction as follows:
\begin{equation}
    (\hat a_{t}, \hat s_{t+1}) \sim DT(R_{t-k+1}, s_{t-k+1}, a_{t-k+1}, ...,R_t, s_t)
\end{equation}
Different from the existing models like GAS \cite{li2025gas} and GAVE \cite{gao2025generative}, we choose to jointly generate the next action $\hat a_{t}$ and the subsequent environment state $\hat s_{t+1}$. This treatment could exploit more supervision signals for DT module training, explicitly guiding the model to capture high-order state evolution. Meanwhile, as described in the following, the estimated $\hat s_{t+1}$ works as the pivot to enable better modeling of transient state transitions by the IDM.

\subsubsection{Inverse Dynamics Module}
After modeling the historical signals, we further incorporate an inverse dynamics module to infer the actions over the transient state transitions. This design provides an alternative pathway for action generation, enhancing the diversity and robustness of the policy.

The IDM operates as follows: given two consecutive environment states $s_t$ and $\hat s_{t+1}$, it estimates the action $\hat a_{t}^{idm}$ that could have led to this state transition. The module is implemented as a neural network $f_{idm}$, typically parameterized as a multilayer perceptron (MLP), which learns the inverse mapping:
\begin{equation}
    \hat a_{t}^{idm} = f_{idm}(s_t, \hat s_{t+1})
\end{equation}
Here, to ensure consistency between training and inference, the input to the inverse dynamics model is the current state $s_t$ and the next state $\hat s_{t+1}$ predicted by DT. During training, $f_{idm}$ is supervised to minimize the mean squared error between the inferred action and the true action recorded in the dataset. Specifically, the loss function for the IDM is calculated as follows:
\begin{equation}
\mathcal{L}_{idm} = \mathbb{E}_{(s_t, a_{t}) \sim \mathcal{D}} \left[ \left\| f_{idm}(s_t, \hat s_{t+1}) - a_{t} \right\|^2 \right]
\end{equation}
where $\mathcal{D}$ represents the training dataset. This objective encourages the IDM to output action predictions that closely align with the actual actions observed in the real trajectories. 

Beyond providing an alternative action source, this design serves a more profound purpose: it implicitly regularizes the state prediction of the Decision Transformer. By tasking the Inverse Dynamics Model with inferring a plausible action from the transition $(s_t, \hat{s}_{t+1})$, we force the DT to generate a future state $\hat{s}_{t+1}$ that is physically reachable from the current state $s_t$. If the DT hallucinates a future state that is inconsistent with the environment's underlying dynamics, the IDM will struggle to reconstruct the correct action, leading to a higher loss that backpropagates to the DT. This feedback loop encourages the DT to learn a more realistic model of environmental evolution, thereby grounding its long-term sequence generation in plausible, moment-to-moment state transitions.

\subsubsection{Two-Stage Training}
Our training procedure adopts a two-stage paradigm to facilitate stable and efficient joint learning of DT and IDM. Specifically, the process is divided into two phases:

\textbf{Phase 1: Separate Training}. During the initial phase, DT and the IDM are optimized independently. Gradients from the IDM are prevented from propagating into the DT by detaching the predicted next state when computing the IDM loss. Given the current state $s_t$ and the detached DT-predicted next state $\hat{s}_{t+1}$, the IDM is trained with the following objective:
\begin{equation}
\mathcal{L}_{idm}' = \mathbb{E}_{(s_t, a_{t}) \sim \mathcal{D}} \left[ \left\| f_{idm}(s_t, \mathrm{stop\_grad}(\hat{s}_{t+1})) - a_{t} \right\|^2 \right]
\end{equation}
where $f_{idm}$ denotes the inverse dynamics model. The DT is updated separately to minimize a behavior cloning loss for actions and a state prediction loss:
\begin{equation}
\mathcal{L}_{dt} = \mathbb{E} \left[ (\hat{a}_t - a_t)^2 + (\hat{s}_{t+1} - s_{t+1})^2 \right]
\end{equation}

\textbf{Phase 2: Joint Training}. After the separate pre-training, both DT and IDM are trained jointly. The IDM loss is incorporated into the total objective for DT, and gradients are allowed to flow through both networks. The training objective becomes:
\begin{equation}
\mathcal{L} = \mathcal{L}_{dt} + \mathcal{L}_{idm}
\end{equation}
In this phase, the IDM takes the non-detached DT-predicted next state $\hat s_{t+1}$ as input.

Compared with single-stage training, this two-phase pipeline prevents unstable gradient propagation in the early stage and enables the IDM to extract high-quality inverse dynamics features. By incorporating inverse dynamics supervision, this approach improves both action generation accuracy and trajectory consistency, helping the DT model learn robust action-state correspondence, especially under complex environment transition scenarios.

\subsection{Q-value-based optimization}

Relying solely on supervised learning from the dataset can only lead to suboptimal behavioral policies, since it restricts exploration and prevents the discovery of better policies. To encourage effective policy exploration within the proposed unified modeling framework, we introduce a Q-value prediction module, implemented as a twin-critic neural network.

\subsubsection{Twin Q Networks and Target Networks Architecture}

Our Q-value estimation module adopts a twin critic architecture, which consists of two independent Q networks, denoted as $Q_1$ and $Q_2$~\cite{fujimoto2021minimalist}. Each network takes a state-action pair $(s_t, a_t)$ as input and outputs the predicted cumulative expected return for taking action $a_t$ in state $s_t$. In addition, each Q network has a corresponding target Q network, namely $Q_1^{\text{target}}$ and $Q_2^{\text{target}}$. The parameters of the target networks are updated via an exponential moving average of the main network parameters, which helps stabilize training. This twin Q network structure alleviates overestimation bias by using $\min(Q_1, Q_2)$ for conservative value estimation and target calculation.

\subsubsection{Critic Training Procedure}

During training, transitions \\ $(s_t, a_t, r_t, s_{t+1}, a_{t+1})$ are sampled from the replay buffer, where $r_t$ is the reward and $d_t$ indicates whether the episode ends. At each update step, we first compute the temporal difference (TD) target $y_t$ using the target networks as follows:
\begin{equation}
y_t = r_t + \gamma (1 - d_t) \min \left\{ Q_1^{\text{target}}(s_{t+1}, a_{t+1}), Q_2^{\text{target}}(s_{t+1}, a_{t+1}) \right\},
\end{equation}
where $\gamma$ is the discount factor. The current Q networks $Q_1$ and $Q_2$ estimate the value for the sampled state-action pairs, and the critic loss is computed as the sum of mean squared errors:
\begin{equation}
\mathcal{L}_{critic} = \mathbb{E} \left[
\left( Q_1(s_t, a_t) - y_t \right)^2 + \left( Q_2(s_t, a_t) - y_t \right)^2
\right].
\end{equation}

\subsubsection{Q-Optimized Actor Training}
By leveraging the critic model, we optimize the training of the actor modules (\ie DT and IDM) through the incorporation of a Q-value regularization term into their loss functions:
\begin{equation}
\mathcal{L}_{actor} = \mathcal{L}_{dt} + \mathcal{L}_{idm} + \mathbb{E}_{s} \left[ -\min(Q_1(s, \hat a), Q_2(s, \hat a)) \right].
\end{equation}
The negative sign in the regularization term incentivizes the actor to generate actions associated with higher Q-values, thereby favoring behaviors that are expected to yield greater returns. This composite loss function strikes a balance between learning behavioral policies from offline data and exploring novel policies with improved performance.

\subsection{Q-value Based Action Selection at Inference}

During inference, our framework enables the generation of two candidate actions, based on both the Decision Transformer and the inverse dynamics module. The Q-value prediction module evaluates each candidate action and selects the one with the highest estimated Q-value:

\begin{equation}
\begin{aligned}
Q_{\text{dt}} &= \min \left\{ Q_1(s, \hat a),\; Q_2(s, \hat a) \right\}, \\
Q_{\text{idm}} &= \min \left\{ Q_1(s, \hat a^{\text{idm}}),\; Q_2(s, \hat a^{\text{idm}}) \right\}, \\
a^* &= \arg\max \left\{ Q_{\text{dt}},\; Q_{\text{idm}} \right\}.
\end{aligned}
\end{equation}

By integrating the Q-value prediction module, \baby provides a principled, value-driven, and flexible decision-making mechanism for selecting among actions estimated by different perspectives, ensuring robust policy deployment and adaptability in dynamic or even unseen advertising environments, as demonstrated by our experimental results.

\subsection{Summary} 
\label{sec:Discussion}
Within the proposed unified modeling architecture, the Q-value optimization module incorporates Q-value regularization into the DT loss, explicitly guiding the model to generate high-value actions and thereby enhancing its exploration capability. Compared with a traditional DT trained solely via behavior cloning, this mechanism is more effective at discovering high-quality out-of-distribution trajectories.  

Therefore, the generated $\hat{a}_t$ could be an effective exploration during the training process, where $\hat{s}_{t+1}$ is the corresponding future state predicted by the DT as well. In these cases, $\hat{a}_t$ would not be equal to the ground truth $a_t$ but leads to a better Q value instead. Note that Q-value module is not involved for IDM training. 
The IDM reconstructs ground-truth actions based solely on the estimated state transition patterns. Hence, the IDM aims to memorize the transition patterns that lead to good explorations generated by the DT. That is to say, the IDM tends to imitate the behavioral policy embedded in the dataset, working as a reliable fallback when exploratory actions may be suboptimal or risky.

\section{Offline Experiment}
\subsection{Experimental Setting}
In this section, we conducted detailed offline experiments to answer the following questions:

\begin{itemize}[left=0pt]
    \item RQ1: Does \baby perform better than other baseline methods across different testing environments?  
    \item RQ2: How does each design choice contribute to the overall performance?  
    \item RQ3: How do DT and IDM co-operate together to improve bidding actions?  
\end{itemize}
We first describe the experimental settings in the following. The code implementation has been released\footnote{https://github.com/M2C-Tech/GUIDE}.

\subsubsection{Datasets}

We use AuctionNet \cite{su2024auctionnet}, a large-scale benchmark proposed by Alibaba, for our evaluation. AuctionNet is the official dataset and simulation environment for the NeurIPS 2024 Advertising Bidding Competition. It is designed to model real-world advertising auto-bidding scenarios and consists of two main components: an \textbf{offline dataset} and a dynamic \textbf{simulation environment}.

The offline dataset simulates competition among 48 different advertisers over multiple advertising periods. Each period contains approximately 500{,}000 impression opportunities and is divided into 48 decision steps. The data is provided in two formats: (1) Traffic-level data, which offers granular records for each impression, including features like predicted conversion probability (pValue), bid, cost, and win status. (2) Trajectory-level data, which aggregates the information into an RL-style format with states, actions, and rewards for each advertiser at each decision step. To ensure our evaluation is rigorous and reflects advanced real-world challenges, we specifically use the \textbf{final-round} AuctionNet dataset, which is characterized by greater data sparsity and higher difficulty compared to the preliminary-round version.

For a more comprehensive and dynamic assessment, we also employ the official simulation environment \cite{su2024auctionnet}. This environment faithfully replicates the multi-agent competitive dynamics of an industrial-scale advertising platform. In each evaluation, our proposed agent controls one advertiser and competes against the other 47 advertisers, which are operated by a diverse set of strong official baseline agents. These baselines span various algorithms, including PID controllers, Online Linear Programming, Offline Reinforcement Learning, and Decision Transformers, creating a heterogeneous and challenging competitive landscape. To ensure fair and robust assessment, the evaluation protocol requires each submitted strategy to sequentially control all 48 advertisers over multiple delivery periods, with the final performance being an aggregation of all results. This two-pronged evaluation approach allows for a thorough and realistic assessment of our method's performance.

\subsubsection{Metrics}
To assess the efficacy of our model, we utilize a performance metric termed the advertising bidding score \cite{su2024auctionnet}. This score quantifies the agent's proficiency in maximizing conversions while strictly adhering to the advertiser's predefined Cost Per Acquisition (CPA) constraint, denoted as $C$. A penalty is applied if the actual CPA surpasses $C$. The score is formally defined as:

\begin{equation}
\label{equ:Score}
Score = \mathbb{P}(CPA;C) \cdot \sum_{i} x_{i} \cdot v_{i}
\end{equation}
Here, the penalty function, which comes into effect when the actual CPA exceeds the constraint $C$, is given by:      
\begin{equation}
\label{equ:P(CPA;C)}
\mathbb{P}(CPA;C) = \min \left\{ \left( \frac{C}{CPA} \right)^{\beta}, \ 1 \right\}
\end{equation}
where $\beta$ is a positive hyperparameter (commonly set to $\beta=2$). This penalty is specifically enforced only when $CPA > C$. This metric enables a robust evaluation of the bidding agent's capacity to balance conversion optimization with cost efficiency.

\subsubsection{Baselines}
To evaluate the performance of our \baby, we compare it against a range of baseline approaches, including both reinforcement learning-based and generative model-based methods. For offline reinforcement learning, we consider \textbf{BC} \cite{bain1995framework}, \textbf{IQL} \cite{kostrikov2021offline}, \textbf{CQL} \cite{kumar2020conservative}, and \textbf{TD3-BC} \cite{fujimoto2021minimalist}, where all RL models are implemented following the AuctionNet \cite{su2024auctionnet} settings. Among generative approaches, we include \textbf{AIGB} \cite{guo2024generative}, a method built upon Decision Diffusion, as well as \textbf{DT} \cite{chen2021decision} and its variants, namely \textbf{GAS}~\cite{li2025gas} and \textbf{GAVE}~\cite{gao2025generative}, all of which are implemented using their released official code. For fairness, the settings of Decision Transformer are kept identical across these generative baselines.

\subsection{RQ1: Offline Dataset and Simulation Environment Evaluation}
As shown in Tables \ref{tab:main} and \ref{tab:sim_env}, we evaluate the performance of different methods on offline datasets under various budget levels, as well as using a simulated environment. Our key findings are as follows:
\begin{itemize}[left=0pt]
    \item \baby outperforms all baselines across all budget levels in offline dataset testing, and also achieves superior performance in simulation environments, demonstrating its strong advantages and adaptability across diverse testing scenarios. By employing unified modeling and trajectory exploration, \baby gains a deeper understanding of the advertising bidding environment, enabling it to maintain high performance across varying conditions.
    \item Generative models with exploration capabilities, such as GAS and GAVE, outperform RL-based models and the basic DT model. This advantage is primarily attributed to their ability to effectively leverage historical bidding information while performing exploration, thereby mitigating the reliance on MDP modeling and overcoming limitations imposed by the dataset behavior policy.
    \item AIGB performed poorly, yielding the worst results in both offline datasets and simulation environments. This may be due to Decision Diffusion's difficulty in effectively learning reasonable policies under conditions of long sequences and sparse rewards. more efficient diffusion architectures combined with appropriate exploration strategies might be necessary for improvement.

\end{itemize}

\begin{table}[htbp]
    \centering
    \caption{Offline evaluation on the AuctionNet under different budgets. Bold indicates the best results, underline denotes the second-best results.}
    \label{tab:main}
    
    \begin{tabular}{lllllll}
        \toprule
        Dataset     & Method & 50\%           & 75\%          & 100\%            & 125\%          & 150\%           \\
        \midrule
        \multirow{9}{*}{AuctionNet}
            & IQL    & 17.9          & 26.9          & 30.9          & 32.0          & 37.8          \\
            & BC     & 15.0          & 20.3          & 26.8          & 31.6          & 36.6          \\
            & CQL    & 16.1          & 22.4          & 27.9          & 32.1          & 37.6          \\
            & TD3-BC & 15.0          & 22.7          & 26.4          & 31.4          & 38.0          \\
            & DT     & 18.4          & 24.9          & 27.6          & 35.6          & 39.4          \\
            & AIGB   & 10.7          & 22.2          & 24.6          & 31.8          & 36.5          \\
            & GAS    & 18.4          & 27.5          & 36.1          & 40.0          & 46.5          \\
            & GAVE   & \underline{19.6} & \underline{28.3} & \underline{37.2} & \underline{42.7} & \underline{47.4} \\
            & \baby    & \textbf{20.3} & \textbf{29.1} & \textbf{37.6} & \textbf{43.3} & \textbf{48.3} \\
        \bottomrule
    \end{tabular}
\end{table}

\begin{table}[htbp]
    \centering
    \caption{Performance in the Simulation Environment}

    \begin{tabular}{
        >{\centering\arraybackslash}m{2cm} 
        >{\centering\arraybackslash}m{2cm} 
        >{\centering\arraybackslash}m{2cm}
    }
        \toprule
        Dataset & Method & Score \\
        \midrule
        \multirow{9}{*}{Simulation Env.} 
            & IQL      & 6534 \\
            & BC       & 6366 \\
            & CQL      & 7138 \\
            & TD3-BC   & 7008 \\
            & DT       & 6920 \\
            & AIGB  & 6248 \\
            & GAS      & \underline{7454} \\
            & \textbf{\baby}     & \textbf{8343} \\
        \bottomrule
    \end{tabular}
    \label{tab:sim_env}
\end{table}

\subsection{RQ2: Model Analysis}\label{ssec:ablation}
\subsubsection{Ablation Study} 
\begin{figure}[htbp]
    \centering
    \includegraphics[width=\linewidth]{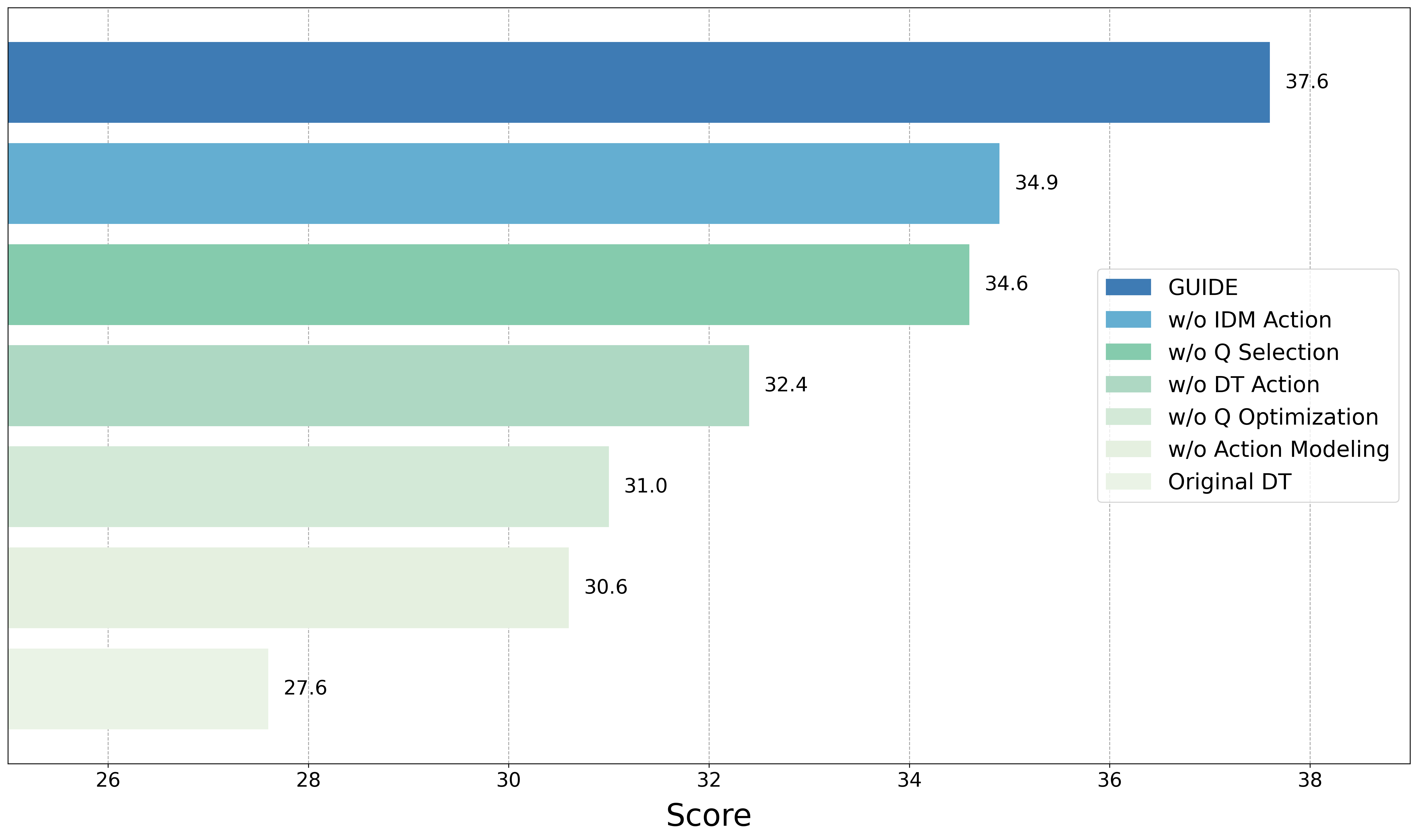}
    \caption{Ablation Study}
    \label{fig:ablation-study}
\end{figure}

As shown in Figure~\ref{fig:ablation-study}, to investigate the effectiveness of each proposed component, we conducted ablation studies by evaluating the following modified versions:
\begin{itemize}[left=0pt]
\item w/o IDM Action: Keep the model structure unchanged, and use only actions from DT.
\item w/o DT Action: Keep the model structure unchanged, and use only actions from IDM.
\item w/o Q Optimization: Only remove Q-value regularization optimization.
\item w/o Q Selection: Retain Q-value regularization optimization, but select actions randomly.
\item w/o action modeling: remove the DT action loss, and use only actions from state modeling.
\item Original DT: Follow the settings of the original DT paper(w/o state modeling), and remove all optimizations.
\end{itemize}

According to Figure~\ref{fig:ablation-study}, we can draw the following findings: First, omitting actions from the DT or IDM in \baby leads to performance degradation, confirming that coupling them together contributes positively and underscoring the necessity of unified modeling.

Second, random selection instead of Q-value-based selection also reduces performance, with results lying between those of using the two action sources separately. This can be attributed to DT actions are higher in overall quality than IDM actions.

Third, removing Q-value regularization optimization causes a significant drop in performance, though still outperforming the original DT. This illustrates the effective role of Q-value regularization, while also highlighting the advantage of the unified modeling architecture compared to the original DT. 

Fourth, removing the action loss and using actions generated by state modeling yields a score only slightly higher than that of the original DT with action modeling, and significantly lower than \baby. This also underscores the importance of joint modeling.

\subsubsection{Two-stage Training Analysis}
\begin{figure}[htbp]
    \centering
    \includegraphics[width=\linewidth]{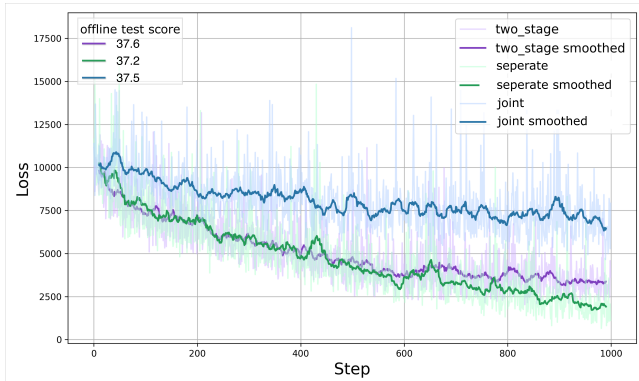}
    \caption{Two-stage Training Analysis}
    \label{fig:two-stage-training}
\end{figure}

We conduct a detailed analysis of the proposed two-stage training strategy. As shown in Figure \ref{fig:two-stage-training}, the blue curve represents joint training of the DT and IDM modules throughout the entire process, where the loss decreases slowly. The purple and green curves correspond to the two-stage training and fully separate training strategies, respectively. In the early stages of training, both purple and green curves are almost identical; in the later stages, after joint training begins, the purple curve’s loss decreases at a slower rate, but still faster than the blue curve. Furthermore, it can be observed that the loss peaks of the purple curve are fewer and lower than those of the blue curve. From the perspective of offline testing scores, the two-stage training strategy also achieves the best performance. Overall, the two-stage training strategy demonstrates both stability and rapid convergence, offering a clear advantage and making it more suitable for practical applications and deployment.

\subsection{RQ3: Co-operation between DT and IDM}
\label{sec:rq3}
\begin{figure}[htbp]
    \centering
    \includegraphics[width=\linewidth]{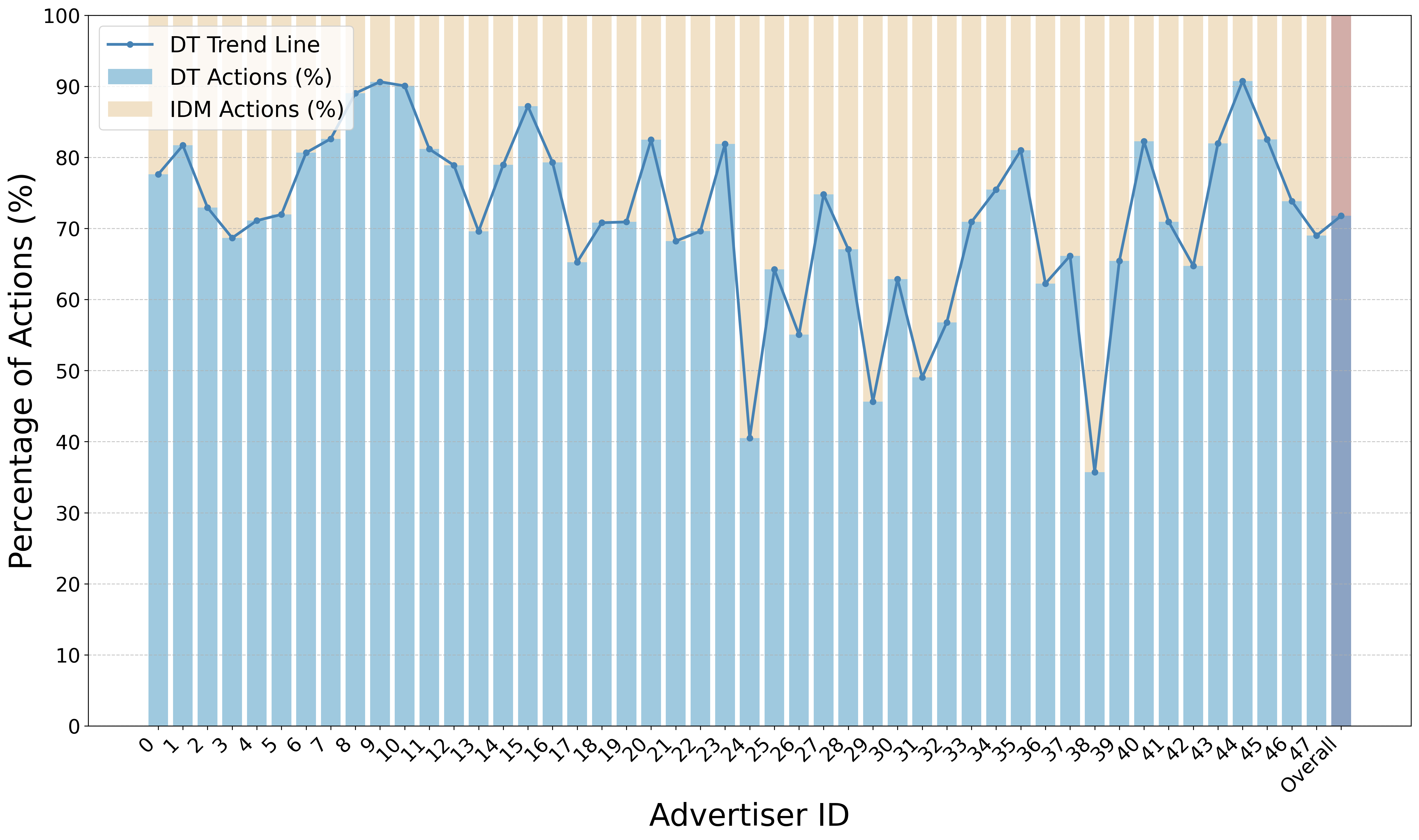}
    \caption{Action preferences of different advertisers}
    \label{fig:action-preferences}
\end{figure}

As shown in Figure \ref{fig:action-preferences}, to investigate the impact of unified modeling on bidding actions, we have statistically analyzed the usage preferences of all 48 advertisers for actions from DT and IDM. Our key findings are as follows:

First, all advertisers utilize actions from both DT and IDM, with no instance of completely ignoring one source, indicating that both channels are effectively integrated within the unified modeling framework and contribute significantly to the bidding process.

Second, the majority of advertisers prefer DT over 70\% of the time, indicating that DT often produces superior actions. This phenomenon is consistent with our model design. That is, by performing explicit action optimization via the Q regularization term, the DT framework can effectively explore OOD trajectories and generally outperform IDM. When DT makes a risky OOD exploration, the model falls back to the more conservative actions generated by IDM as a safeguard. We achieve a unified balance between exploration and safety.

Third, we conducted a detailed analysis of certain advertisers who show a preference for IDM to investigate the causes of this phenomenon. To do this, we classified advertisers based on budget and CPA constraint levels. Specifically, advertisers were grouped into three budget tiers: high budget (top 30\%), medium budget (middle 40\%), and low budget (bottom 30\%). Similarly, CPA constraints were classified into high constraint (top 30\%), medium constraint (middle 40\%), and low constraint (bottom 30\%).

We conducted multiple experiments and averaged the results to reduce statistical noise. The analysis identified four advertisers (No.\ 24, No.\ 29, No.\ 31, and No.\ 38) who consistently exhibited a higher preference for IDM-generated actions. All four belonged to one of the following extreme budget--constraint configurations:
\begin{itemize}
    \item High budget combined with low constraint
    \item Low budget combined with high constraint
\end{itemize}
These findings suggest that an advertiser's budget and constraint settings can significantly influence the preference between DT and IDM strategies. In particular, we believe that such extreme cases of misalignment between budget and CPA constraints can lead to errors in DT during exploration, causing the model to prefer the more conservative IDM policy.

\begin{figure}
    \centering
    \includegraphics[width=\linewidth]{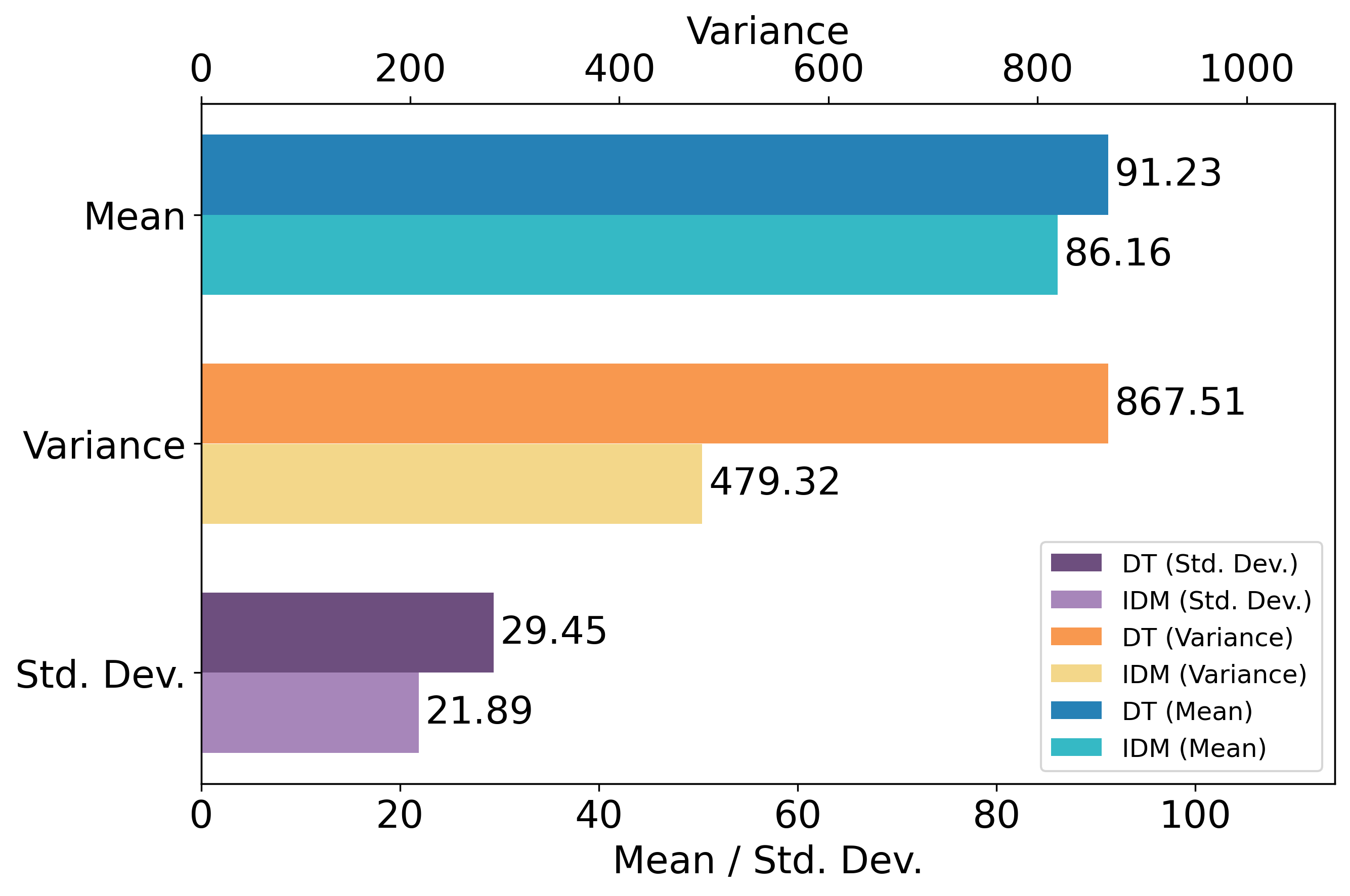}
    \caption{Volatility comparison between DT and IDM bid actions}
    \label{fig:action_volatility}
\end{figure}

In addition, we analyzed the volatility characteristics of actions generated by DT and IDM to further explain IDM's role as a safeguard. Figure~\ref{fig:action_volatility} shows the mean, variance, and standard deviation of bid actions for both methods. These results quantitatively confirm that the overall volatility of bidding actions generated by IDM is smaller than its counterpart from DT. This is consistent with DT’s exploratory nature, as discussed in Section~\ref{sec:Discussion}, and reinforces IDM's function as a more stable and conservative safeguard within the unified model.

\section{Online A/B Test}

\subsection{Deployment}

To assess the real-world performance of \baby, we deployed it on Taobao, a major e-commerce platform in China. We use the DT model as the baseline for comparison with \baby. This setting involves advertisers specifying their budgets and optionally setting various constraints, such as CPA or Return on Investment (ROI). The bidding policy is responsible for optimizing traffic value while strictly adhering to these constraints.

\textbf{State Representation}:
Each campaign’s bidding environment is described by a 19-dimensional state vector comprising campaign-specific indicators (e.g., fraction of budget and bidding steps remaining, deviation from target CPA) and temporal statistics aggregated over recent time steps. Features include impressions, clicks, conversions, ad cost, GMV, and derived metrics like CTR and CVR, offering a detailed snapshot of the campaign’s progress and recent market dynamics.

\textbf{Action Mechanism}:
Rather than apply the model’s action output directly, we stabilize bid adjustments by smoothing the action with a windowed average over the preceding time steps. Concretely, the final bid coefficient at each decision point is computed by blending the model’s current suggestion with the average of previous coefficients over a trailing two-hour window. This technique helps mitigate abrupt fluctuations in bidding strategies, promoting more consistent campaign outcomes.

\textbf{Reward and Return-to-Go Formulation}:
The reward signal is grounded in the advertiser's key business objective: maximizing gross merchandise value (GMV), subject to CPA and budget constraints. Specifically, the return-to-go is defined as the expected cumulative GMV from the current time frame through the remainder of the promotional day, conditioned on considering all specified limits and penalty term. 

\textbf{Evaluation Metrics}:
The effectiveness of the deployed \baby policy is assessed using multiple key business and platform-level metrics. These include:
\begin{itemize}[left=0pt]
    \item \textit{Ad Click}: Number of clicks on ads, reflecting user engagement with advertised content.
    \item \textit{Ad Cost}: Total advertising expenditure incurred via bidding, representing the financial outlay for acquiring traffic.
    \item \textit{Ad GMV}: Gross Merchandise Volume generated from ad clicks, measuring the transaction value directly attributable to advertising.
    \item \textit{Ad ROI}: Return on Investment, calculated as the ratio of Ad GMV to Ad Cost, indicating the efficiency and profitability of ad cost.
\end{itemize}

\paratitle{Training Phase.} Our online system is trained on one week of historical advertising campaign data, where each trajectory represents a single day’s sequence of observed states, decisions, and outcomes. The core model uses a Decision Transformer architecture with 6 stacked Transformer blocks, each featuring 8 attention heads and a hidden dimension of 512. The multilayer perceptrons in both the Inverse Dynamics Model and Q modules have a hidden dimension of 256.

\paratitle{Inference Phase.} Deployed on a leading ad platform, \baby serves as the bidding agent for all advertised items, generating and updating bid decisions every 30 minutes across the entire item set to enable dynamic ad allocation aligned with real-time market conditions and user intent. A large-scale online A/B test covered approximately 160,000 distinct products and impacted tens of millions of dollars in gross merchandise value (exact figure withheld per company policy), consistently demonstrating significant gains in both efficiency and effectiveness, fully validating the system’s robustness and business value.

\subsection{Online A/B Test Results}
\subsubsection{Overall Performance}
Table \ref{tab:Improvements of GUIDE on Key Advertising Metrics} shows significant improvements achieved by \baby on key advertising metrics. During an 8-day online A/B test, Ad Clicks increased by 1.40\%, Ad Cost increased by 1.66\%, Ad GMV increased by 4.10\%, and Ad ROI increased by 3.52\%. These performance gains fully demonstrate the effectiveness of \baby in real-world applications.

It is particularly insightful to analyze the interplay between these metrics. The modest increase in Ad Cost (1.66\%) accompanied by a substantially larger increase in Ad GMV (4.10\%) is a strong indicator of improved spending efficiency. This is not merely a case of bidding more aggressively to gain more traffic; rather, it demonstrates that \baby allocates the budget more intelligently. The resulting significant lift in Ad ROI (+3.52\%) confirms that each dollar spent under \baby's control is generating higher returns. This demonstrates that the model’s unified approach allows it to identify and secure higher-value ad impressions, specifically those with a greater likelihood of conversion, while steering clear of wasteful spending on less promising opportunities. As a result, the performance improvements stem not just from increased traffic volume, but more importantly from the enhanced quality and profitability of the acquired traffic, which is the ultimate objective of any advanced auto-bidding system.

\begin{table}[htbp]
    \centering
    \caption{Improvements of GUIDE on Key Advertising Metrics}
    \label{tab:Improvements of GUIDE on Key Advertising Metrics}
    \begin{tabular}{ccccc}
        \toprule
        Metric & Ad Click & Ad Cost & Ad GMV & Ad ROI \\
        \midrule
        Improvement & 1.40\% & 1.66\% & 4.10\% & 3.52\% \\
        \bottomrule
    \end{tabular}
\end{table}
\subsubsection{Trajectory optimization capability}

A foundational principle in computational advertising is that advertising revenue is maximized when the expenditure trend dynamically aligns with the natural fluctuations in user traffic~\cite{agarwal2014budget}. To evaluate \baby’s ability to perform such dynamic budget allocation, we conducted an analysis of its cost trajectory control. We began by constructing an ideal ad cost trajectory in a post-hoc manner, which serves as an oracle benchmark representing the optimal spending pattern proportional to the observed traffic.

\begin{figure}[H]
    \centering
    \includegraphics[width=1.1\linewidth]{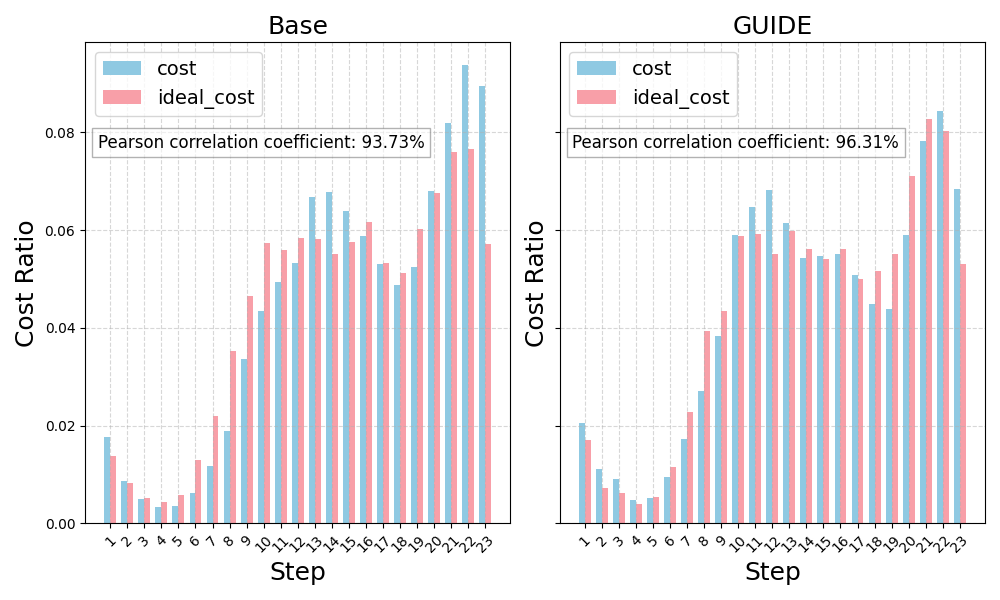}
    \caption{Cost Trajectory Analysis}
    \label{fig:The Cost Trajectory for Base and GUIDE}
\end{figure}

Figure \ref{fig:The Cost Trajectory for Base and GUIDE} visually compares the actual cost trajectories managed by a baseline method and our proposed \baby against this ideal trajectory. As can be seen, while both methods attempt to follow the general trend, the cost distribution controlled by \baby (right panel) tracks the ideal cost much more closely across the different time steps. The baseline method (left panel), in contrast, shows more significant deviations, particularly exhibiting over-expenditure during several peak hours. To quantify this alignment, we employed the Pearson correlation coefficient. The trajectory controlled by \baby achieved a correlation of 96.31\% with the optimal trajectory, which is a clear improvement over the baseline's 93.73\%. This result quantitatively confirms that \baby possesses superior trajectory optimization capabilities, enabling more precise and effective budget pacing over time.

\section{Conclusion and Limitations}

We present \baby, a unified modeling and exploration approach for automatic Ad bidding. \baby integrates three carefully designed components: the Decision Transformer (DT), the Inverse Dynamics Model (IDM), and the Q-value module. Their synergistic interaction enables a balance between exploration and safety, leading to significant improvements in bidding strategies. Experiments show that \baby consistently outperforms baseline methods across offline data, simulations, and real-world applications. In online tests, \baby increased Ad GMV by 4.10\%,  providing an effective solution for automatic bidding in complex advertising scenarios.

Despite \baby’s strong performance in ad bidding, it lacks fine-grained mechanisms to handle abrupt traffic changes, limiting its responsiveness during sudden fluctuations or special events. Moreover, it primarily relies on offline data and current model architectures. Future work could integrate LLMs for trajectory control and dynamic optimization, enhancing robustness and adaptability in evolving marketplaces.

\begin{acks}
This work was supported by Alibaba Group through Alibaba Innovative Research Program; and it was also supported by National Natural Science Foundation of China (No. 62272349).
\end{acks}

\bibliographystyle{ACM-Reference-Format}
\bibliography{sample-base.bib}

\end{document}